\documentclass[10pt,twocolumn,letterpaper]{article}

\usepackage{iccv}
\usepackage{times}
\usepackage{epsfig}
\usepackage{graphicx}
\usepackage{amsmath}
\usepackage{amssymb}

\usepackage{multirow}
\usepackage{hhline}
\usepackage{adjustbox}
\usepackage{paralist}
\usepackage{tabularx}
\usepackage[pagebackref=true,breaklinks=true,letterpaper=true,colorlinks,bookmarks=false]{hyperref}

\iccvfinalcopy 

\ificcvfinal\fi

\begin{document}

\title{WaveFill: A Wavelet-based Generation Network for Image Inpainting}

\author{Yingchen Yu$^{1}$\quad \quad
Fangneng Zhan$^1$\quad \quad
Shijian Lu$^1$\thanks{Corresponding author}\quad \quad
Jianxiong Pan$^2$\\
Feiying Ma$^2$\quad \quad
Xuansong Xie$^2$\quad \quad
Chunyan Miao$^1$\\
$^1$ Nanyang Technological University $\ \
^2$ DAMO Academy, Alibaba Group\\
{\tt\small yingchen001@e.ntu.edu.sg, \{fnzhan, shijian.lu, ascymiao\}@ntu.edu.sg} \\
{\tt\small \{jianxiong.pjx, feiying.mfy\}@alibaba-inc.com, xingtong.xxs@taobao.com} 
}

\maketitle
\ificcvfinal\thispagestyle{empty}\fi

\begin{abstract}
Image inpainting aims to complete the missing or corrupted regions of images with realistic contents.
The prevalent approaches adopt a hybrid objective of reconstruction and perceptual quality by using generative adversarial networks. However, the reconstruction loss and adversarial loss focus on synthesizing contents of different frequencies and simply applying them together often leads to inter-frequency conflicts and compromised inpainting. This paper presents WaveFill, a wavelet-based inpainting network that decomposes images into multiple frequency bands and fills the missing regions in each frequency band separately and explicitly. WaveFill decomposes images by using discrete wavelet transform (DWT) that preserves spatial information naturally. It applies L1 reconstruction loss to the decomposed low-frequency bands and adversarial loss to high-frequency bands, hence effectively mitigate inter-frequency conflicts while completing images in spatial domain. 
To address the inpainting inconsistency in different frequency bands and fuse features with distinct statistics, we design a novel normalization scheme that aligns and fuses the multi-frequency features effectively. Extensive experiments over multiple datasets show that WaveFill achieves superior image inpainting qualitatively and quantitatively.
\end{abstract}
\section{Introduction}

As an ill-posed problem, image inpainting 
is not to recover the original images for corrupted regions but to synthesize alternative contents that are visually plausible and semantically reasonable. It has been widely investigated in various image editing tasks such as object removal, old photo restoration, movie restoration, and so on. Realistic and high-fidelity image inpainting remains a challenging task especially when the corrupted regions are large and have complex texture and structural patterns. 
\begin{figure}[t]
\centering
\includegraphics[width=1.0\linewidth]{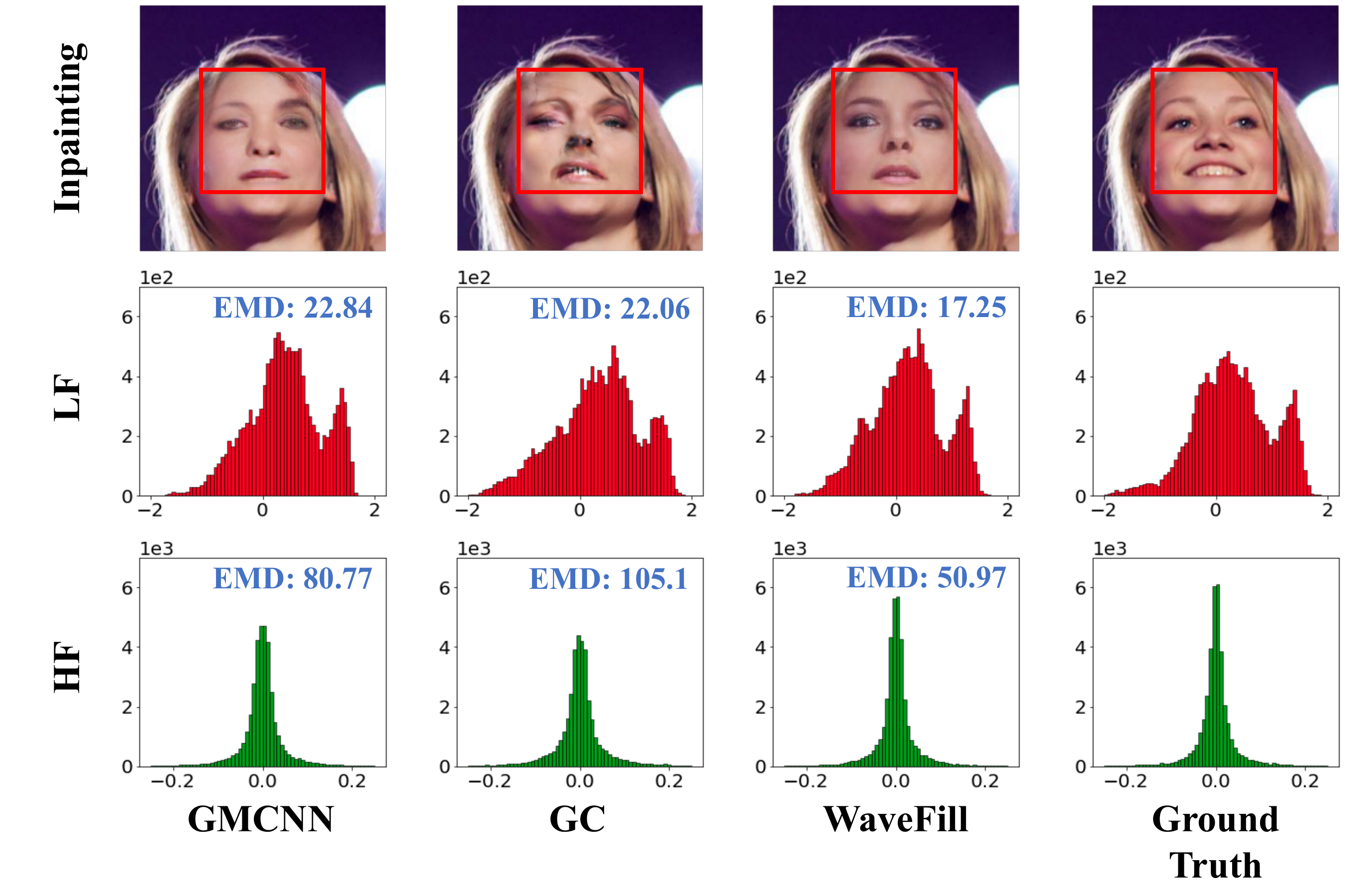}
\caption{ 
Image inpainting often faces a dilemma of reconstruction and perceptual quality: L1/L2 loss focuses on the reconstruction of global low-frequency structures while adversarial loss focuses on generating high-frequency texture details. State-of-the-art approaches implicitly tackle this issue by weighted summing of the two objectives (e.g. in GMCNN \cite{wang2018image}) or employing a Coarse-to-Fine strategy (e.g. in GC \cite{yu2019free}), but tend to produce inconsistent distributions with missing details or artifacts. The proposed \textit{WaveFill} disentangles images into multiple frequency bands and applies relevant losses to different bands separately, which mitigates inter-frequency conflicts and produces more realistic structures and details. The distances between the ground-truth histogram and prediction histograms in both low-frequency (LF) and high-frequency (HF) are evaluated by Earth Mover's Distance (EMD) \cite{rubner2000earth}.
}
\label{fig_intro}
\end{figure}

State-of-the-art image inpainting methods leverage generative adversarial networks (GANs) \cite{goodfellow2014generative} heavily for generating realistic high-frequency details \cite{pathak2016context}.  But they often face a dilemma of perceptual quality and reconstruction that share a \textit{perception-distortion trade-off} \cite{blau2018perception}. Specifically, the adversarial loss in GANs tends to recover high-frequency texture details and improve the perceptual quality \cite{sajjadi2017enhancenet, deng2019wavelet}, while the L1/L2 loss in reconstruction focuses more on recovering low-frequency global structures \cite{pathak2016context}. Concurrently optimizing the two objectives in spatial domain tends to introduce inter-frequency conflicts as illustrated in Fig.~\ref{fig_intro}. 
GMCNN \cite{wang2018image} balances the two objectives by weighted sum, but it still works in spatial domain with mixed frequency and struggles to generate more realistic high-frequency details due to the inter-frequency conflicts. Gate Convolution (GC) \cite{yu2019free} mitigates this issue by adopting a Coarse-to-Fine strategy \cite{yu2018generative, song2018contextual, ren2019structureflow, liu2018image,yu2019free} that first predicts global low-frequency structures and then refines high-frequency texture details. The coarse estimation network is generally trained with L1 loss, but the inter-frequency conflicts still exist in the refinement network. Moreover, the two-stage network often suffers from inconsistency in generated structure and texture details due to the lack of effective alignment and fusion of multi-stage features \cite{Liu2019MEDFE}.
To address the aforementioned issues, we design WaveFill, an innovative image inpainting framework that employs wavelet transform to complete corrupted image regions at multiple frequency bands separately. Specifically, we convert images into wavelet domain with 2D discrete wavelet transform (DWT) \cite{cotter2020uses} where the images can be disentangled into multiple frequency bands accurately without losing spatial information. The disentanglement allows us to apply adversarial (or L1) loss to the high-frequency (or low-frequency) branches explicitly and separately, which greatly mitigates the content conflicts as introduced by concurrently optimizing the two different objectives over entangled features in spatial space. In addition, we design a novel \textit{frequency region attentive normalization (FRAN)} scheme that aggregates attention from low frequency to high frequency to align and fuse the multi-frequency features. FRAN ensures the consistency across multiple frequency bands and helps suppress artifacts and preserve texture details effectively. The separately completed features in different frequency bands are then transformed back to the spatial domain via inverse discrete wavelet transform (IDWT) to produce the final completion.

The contributions of this work can be summarized in three aspects. First, we propose WaveFill, an innovative image inpainting technique that synthesizes corrupted image regions at different frequency bands explicitly and separately, which effectively mitigates the inter-frequency conflicts while minimizing adversarial and reconstruction losses. Second, we design a novel normalization scheme that enables attentive alignment and fusion of the multi-frequency features with effective artifact suppression and detail preservation. Third, extensive experiments over multiple datasets show that the proposed WaveFill achieves superior inpainting as compared with the state-of-the-art.
\begin{figure*}[t]
\centering
\includegraphics[width=\linewidth]{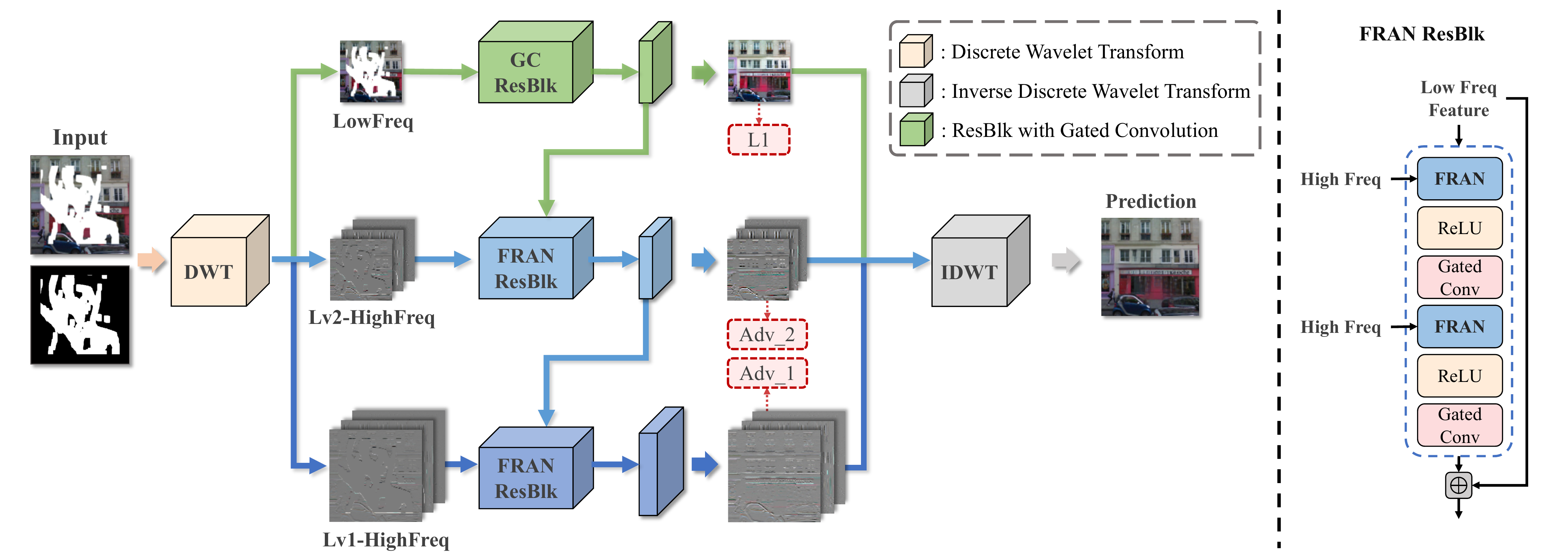}
\caption{
The architecture of the proposed WaveFill inpainting network: The WaveFill generator consists of three branches for processing information of different frequencies separately. Given an \textit{Input Image}, we first decompose it into multiple frequency bands via \textit{DWT} (discrete wavelet transform) and then assemble the decomposed frequency bands into three `broadbands’ \textit{LowFreq}, \textit{Lv2-HighFreq} and \textit{Lv1-HighFreq}. The completion is performed in the \textit{LowFreq} first with \textit{GC ResBlk} and the generated features are then aligned and propagated to high-frequency branches (via a novel normalization scheme \textit{FRAN}) for further completion. The L1 loss is explicitly applied in the low-frequency branch, and two discriminators are employed to enable adversarial training in the two high-frequency branches. The generated features in the three branches are finally transformed back to spatial domain via \textit{IDWT} (inverse DWT) to produce the final \textit{Prediction}.
}
\label{fig_arch}
\end{figure*}

\section{Related Works}
\subsection{Image Inpainting}
Image inpainting has been studied for years and earlier works employ diffusion and image patches heavily. Specifically, diffusion methods \cite{bertalmio2000image, ballester2001filling} propagate neighboring information towards the corrupted regions but often fail to recover meaningful structures with little global information. Patch-based methods \cite{barnes2009patchmatch,darabi2012image} complete images by searching and transferring similar patches from the background. They work well for stationary texture but struggle while generating meaningful semantics for non-stationary data.

With the recent advance of deep learning, deep neural networks have been widely explored for the image generation and inpainting \cite{zhan2019sfgan,zhan2021unite,zhan2021rabit,wu2020leed,wu2020cascade,yu2021diverse}. 
In particular, generative adversarial networks \cite{goodfellow2014generative} have been developed to complete images with both faithful structures and plausible appearance. For example, 
Pathak et al. \cite{pathak2016context} proposes a GAN-based method to complete large corrupted regions. 
Yu et al. \cite{yu2018generative} adopts the patch match idea and introduces a Coarse-to-Fine strategy to recover low-frequency structures and high-frequency details. 
Nazeri et al. \cite{nazeri2019edgeconnect} introduces EdgeConnect to predict salient edges without coarse estimation.
Wang et al. \cite{wang2018image} employs multiple branches with different receptive fields for inpainting. 
Liu et al. \cite{Liu2019MEDFE} recovers structures and textures by representing them with deep and shallow features. 
Liu et al. \cite{liu2018image} introduces partial convolution with free-from masks for inpainting. On top of it, Yu et al. \cite{yu2019free} presents gated convolution for inpainting. 

Though the aforementioned methods address image completion in different manners, most of them work in the spatial domain where information of different frequencies is mixed and often introduces inter-frequency conflicts in learning and optimization. Our method instead decomposes images into the frequency space and applies different objectives to different frequency bands explicitly and separately, which mitigates inter-frequency conflicts and improves image inpainting quality effectively. 

\subsection{Wavelet-based Methods}
Wavelet transforms decompose a signal into different frequency components and has shown great effectiveness in various image processing tasks \cite{mallat1999wavelet}. Wavelet-based inpainting has been investigated far before the prevalence of deep learning. For example, Chan et al. \cite{chan2006total} designs variational models with total variation (TV) minimization for image inpainting, and it's improved in \cite{zhang2010wavelet} with non-local TV regularization. 
In addition, Dobrosotskaya et al. \cite{dobrosotskaya2008wavelet} combines diffusion with the non-locality of wavelets for better sharpness in inpainting. 
Zhang and Dai \cite{zhang2012image} decomposes images in the wavelet domain to generate structures and texture with diffusion and exemplar-based methods, respectively. 
The aforementioned methods leverage hand-crafted features which cannot generate meaningful content for large corrupted regions. We borrow the idea of wavelet-based decomposition and incorporate CNN representations and adversarial learning which mitigates this issue effectively.

Recently, incorporating wavelets into deep networks has been explored in various computer vision tasks such as super-resolution \cite{huang2017wavelet,deng2019wavelet}, style transfer \cite{liu2019attribute}, quality enhancement \cite{wang2020multilevel} and image demoir{\'e}ing \cite{liu2020wavelet}. Different from them that directly concatenate frequency bands and pass them to convolutional layers, we design separate network branches to explicitly generate contents for each group of frequency bands, and meanwhile incorporate features from other branches for better completion.

\section{Proposed Method}
\subsection{Overview}
The overview of our proposed inpainting network is illustrated in Fig. \ref{fig_arch}. An input image is first decomposed and assembled into 3 frequency bands \textit{LowFreq}, \textit{Lv2HighFreq} and \textit{Lv1HighFreq}, which are then fed to three network branches for respective completion. We apply L1 reconstruction loss to \textit{LowFreq} and adversarial loss to \textit{Lv2HighFreq} and \textit{Lv1HighFreq} to mitigate the inter-frequency conflicts. In addition, we design a novel normalization scheme \textit{FRAN} that aligns and fuses features from the three branches to enforce the completion consistency across the three frequency bands. The generation results in the three branches are finally transformed back to the spatial domain to complete the inpainting, more details to be described in the ensuing subsections. 

\subsection{Wavelet Decomposition}
The key innovation of our work is to disentangle images into multiple frequency bands and complete the images in different bands separately in the wavelet domain. We adopt 2D discrete wavelet transform (DWT) to first decompose images into multiple wavelet sub-bands with different frequency contents. For each iteration of the decomposition, the DWT applies low-pass and high-pass wavelet filters alternatively along image columns and rows (followed by downsampling), which produces 4 sub-bands including $LL, LH, HL$, and $HH$. The decomposition continues iteratively on $LL^{n-1}$ to produce $LL^n, LH^n, HL^n$, and $HH^n$ until the target level of decomposition $N_w$ is reached. Hence, a total number of $3N_w+1$ wavelet sub-bands will be finally produced including  $LL^{N_w}, \{LH^n\}_{n=1}^{N_w},\{HL^n\}_{n=1}^{N_w}$, and $\{HH^n\}_{n=1}^{N_w}$.
Here $LL^{N_w}$ captures low-frequency information at the $N_w$-th level, $LH^n$, $HL^n$ and $HH^n$ capture the horizontal, vertical and diagonal high-frequency information at the $n$-th level, respectively. Note that the sizes of sub-bands at the $n$-th level are down-sampled with a factor of $1/2^n$. 

In this work, we adopt the Haar wavelet filter as the basis for the wavelet transform, where the high-pass filter is $h_{high} = (1/\sqrt{2},1/\sqrt{2})$ and the low-pass filter is $h_{low} = (1/\sqrt{2},-1/\sqrt{2})$. The level of wavelet decomposition ${N_w}$ is empirically set to 2, we treat $LL^2$ as low-frequency, concatenate $LH^n$, $HL^n$ and $HH^n$ in the channel dimension as $n$-th level high-frequency. Given a input image of size $H \times W \times 3$, we will thus obtain three inputs in the wavelet domain, namely, \textit{LowFreq} with size of $H/4 \times W/4 \times 3$, \textit{Lv2-HighFreq} with size of $H/4 \times W/4 \times 9$ and \textit{Lv1-HighFreq} with size of $H/2 \times W/2 \times 9$. 


\subsection{Frequency Region Attentive Normalization}
It is a vital step to align and fuse the low-frequency and high-frequency features for generating consistent and realistic contents across different frequency bands.
An effective fusion of low-frequency and high-frequency features has two major challenges. First, the statistics of low-frequency and high-frequency bands have clear differences, direct summing or concatenating them could greatly suppress high-frequency information due to its high sparsity. 
Second, the different branches are trained with their explicit loss terms, and the learning capacity (No. of CNN layers and kernel sizes) also varies among the branches. Thus, when inpainting different branches independently without inter-branch alignment, a network branch may generate contents that are reasonable in its own frequency bands but inconsistent across frequency bands of other branches (in object shapes or sizes).
Both issues could lead to various blurs and artifacts in the completion results. We design a novel Frequency Region Attentive Normalization (FRAN) technique that aligns and fuses low-frequency and high-frequency features for more realistic inpainting.

For the issue with the statistical difference, we propose to align the low-frequency features with the target high-frequency features so as to fuse them effectively and alleviate the difficulty of generating target high-frequency bands. Inspired by the spatially-adaptive normalization (SPADE) \cite{park2019semantic}, we achieve the feature alignment by injecting the learnable modulation parameters $\boldsymbol{\gamma}_{H}$ and $\boldsymbol{\beta}_{H}$ of high-frequency features $X_H = \{x_H^1,..., x_H^N\}$ to the low-frequency features $X_L = \{x_L^1,..., x_L^N\}$, where $N$ is the number of spatial positions, i.e. $N = H \times W$.

\begin{figure}[t]
\centering
\includegraphics[width=1.0\linewidth]{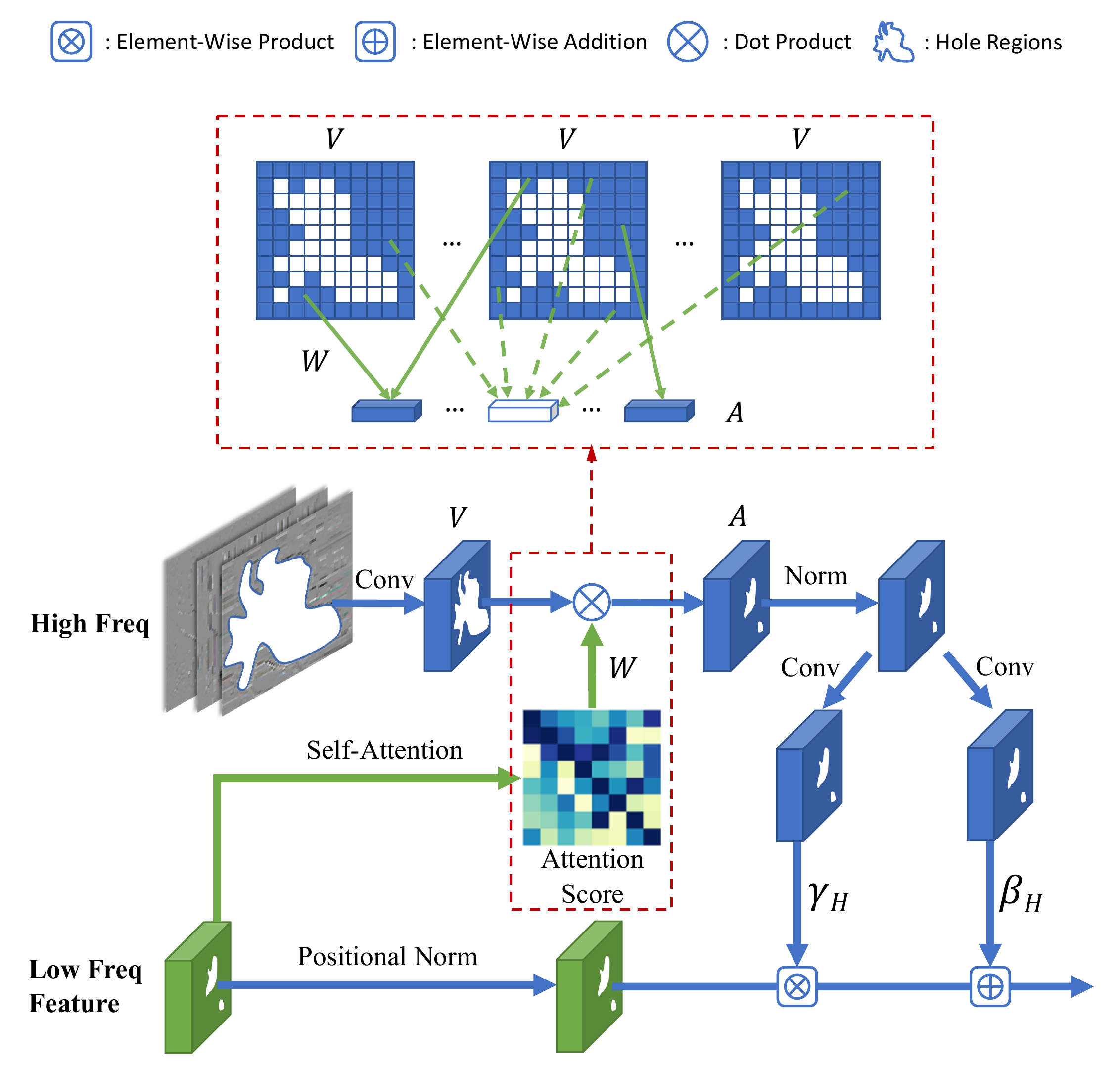}
\caption{
The structure of the proposed Frequency Region Attentive Normalization (FRAN): The irregular regions in each feature map denote corrupted regions. After projecting high-frequency information to a feature space, FRAN aligns the corrupted regions of low-frequency and high-frequency features by aggregating the attention score of low-frequency to high-frequency. The aligned high-frequency features are then convolved to produce modulation parameters $\boldsymbol{\gamma}_{H}$ and $\boldsymbol{\beta}_{H}$ that are injected into the normalized low-frequency features.
}
\label{fig_norm}
\end{figure}

To align the contents in the missing regions, we aggregate the self-attention score of low-frequency features to high-frequency features. Since the attention map depicts the correlation between low-frequency feature patches, the misaligned high-frequency features of corrupted regions can be reconstructed by collectively aggregating features from uncorrupted regions. Another advantage of applying attention aggregation is to leverage complementary features of distant regions by establishing long-range dependencies. As shown in Fig. \ref{fig_norm}, the attention scores $W_{j,i}$ are computed from low-frequency features $X_L \in R^{C\times{N}}$ ($C$ is the channel number) which are firstly transformed to two features space for key and query respectively, i.e. $K = f(X_L), Q = g(X_L)$, $f$ and $g$ are the $1\times{1}$ convolutions. For efficiency, we employ max-pooling to obtain a spatial dimension of $N=1024$ $(32\times{32})$ for attention calculation and aggregation.
\begin{small}
\begin{equation}
    W_{j,i} = \frac{\exp(s_{i,j})}{\sum_{i=1}^N{\exp({s_{i,j})}}},\text{ where } s_{i,j} = f(x_{L}^i)^Tg(x_{L}^j) \ .
\end{equation}
\end{small}

The high-frequency features $X_H$ is then mapped to the feature space with the same hidden dimension by $V = h(X_H)$ where $h$ is the transformation function by convolution. The aggregation of $X_H$ at position $i$ is defined by:
\begin{equation}
    A_{i} = \sum_{j=1}^N{W_{j,i}h(x_{H}^i)}\ .
\end{equation}

Since the high-frequency features are significantly sparse, the magnitude of resultant aggregation is relatively small. We adopt a parameter-free positional normalization \cite{li2019positional} to normalize it and meanwhile preserve structure information. The same normalization is also applied to low-frequency features before the modulation. Finally, the aggregation output $A$ is convolved to produce the modulation parameters $\boldsymbol{\gamma}_{H}$ and $\boldsymbol{\beta}_{H}$ to modulate the normalized low-frequency features:
\begin{equation}
    H = \boldsymbol{\gamma}_{H} \frac{X_L - \mu_{L}}{\sigma_{L}} + \boldsymbol{\beta}_{H} \ ,
\end{equation}
where $H$ is the modulated features, $\mu_{L}$ and $\sigma_{L}$ is the mean and standard deviation of $X_L$ along the channel dimension.

\subsection{Network Architecture}
Our network consists of one generator and 2 discriminator as illustrated in Fig. \ref{fig_arch}.

\textbf{Generation Network.} The generation network consists of 3 branches \textit{LowFreq}, \textit{Lv2-HighFreq} and \textit{Lv1-HighFreq} that recover corrupted regions separately. The \textit{LowFreq} branch consists of a completion module \textit{GC ResBlk} that adopts gated convolution \cite{yu2019free} and residual connection \cite{he2016deep}. Specifically, \textit{GC ResBlk} consists of several consecutive residual blocks with growing dilation rates up to 16 to increase the receptive field. Meanwhile, it replaces all convolutions by gated convolution to dynamically handle missing regions. The generated low-frequency features will be propagated to a decoder that has two gated convolutions to predict the completion of low-frequency sub-bands. Besides, they will also be transferred to two high-frequency branches for guiding and aligning with their generation. 

The high-frequency branch \textit{Lv2-HighFreq} consists of a new residual block \textit{FRAN ResBlk} that is introduced with FRAN as illustrated in Fig. \ref{fig_arch} (right). As the learned modulation parameters have encoded high-frequency information, we directly feed the high-frequency bands to the FRAN without additional encoding. After injecting the high-frequency information to low-frequency features, we propagate the acquired high-frequency features to a separate decoder which also consists of two gated convolutions. Another high-frequency branch \textit{Lv1-HighFreq} shares similar structures with \textit{Lv2-HighFreq}, except that it concatenates the well-aligned and normalized features from the previous two branches and up-sampling them to the current spatial dimension. The generation network thus predicts the inpainting of all 3 frequency bands, and finally converts them back to the spatial domain via inverse Discrete Wavelet Transform (IDWT). As DWT and IDWT are both differentiable, the network can be trained end-to-end.

\textbf{Discrimination Network.} To synthesize high-frequency information, we adopt two discriminators of the same structure to predice \textit{Lv2-HighFreq} and \textit{Lv1-HighFreq}, respectively. Motivated by PatchGAN \cite{isola2017image} and global and local GANs \cite{iizuka2017globally}, we adopt global and local sub-networks on top of PatchGAN to ensure the generation consistency. Additionally, we append a self-attention layer \cite{zhang2019self} after the last convolutional layer to assess the global structure and enforce the geometric consistency.

\subsection{Loss Functions}
We denote the finally completed image by $I_{out}$, the predictions in the wavelet domain by $\{L_{out}^{N_w}, H_{out}^1, ..., H_{out}^{N_w}\}$ ($N_w$ is number of levels in wavelet decomposition), the ground-truth image by $I_{gt}$ and its corresponding wavelet coefficients by $\{L_{gt}^{N_w}, H_{gt}^1, ..., H_{gt}^{N_w}\}$. $D_{n}$ is the discriminator for the $n$-th level high-frequency wavelet coefficients in the wavelet domain.

\textbf{Low-Frequency L1 Loss.} We explicitly employ the L1 loss on the low-frequency subbands in the wavelet domain, which can be defined by:
\begin{align}
    \mathcal{L}_{LF} = ||L_{out}^{N_w} - L_{gt}^{N_w}||_{1} \ .
\end{align}

\textbf{Adversarial Loss.}
For the 2 discriminators of high-frequency branches, we apply the same adversarial losses to them using hinge loss \cite{isola2017image}.
The adversarial loss for discriminator $D_n$ is defined as:
\begin{equation}
\begin{split}
    \mathcal{L}_{D_n} = & \ \mathbb{E}_{H_{gt}^n}[\textit{ReLU}(1-D_n(H_{gt}^n)] \\
                        & + \mathbb{E}_{H_{out}^n}[\textit{ReLU}(1+D_n(H_{out}^n)] \ .
\end{split}
\end{equation}
For the generator, we sum up the adversarial loss of each discriminator to obtain the final loss $\mathcal{L}_{G}$ as below: 
\begin{equation}
\begin{split}
    \mathcal{L}_{G} =  - \sum_{n=1}^{N_w}\mathbb{E}_{H_{out}^n}[D_n(H_{out}^n)] \ .
\end{split}
\end{equation}

\textbf{Feature Matching Loss.}
As the training could be unstable due to the sparsity of high-frequency bands, we adopt the feature matching loss following pix2pixHD \cite{wang2018high} on both the two discriminators to stabilize the training process.
\begin{equation}
\begin{split}
    \mathcal{L}_{FM} = \sum_{n=1}^{N_w}\mathbb{E}\left[\sum_{i=1}^L{\frac{1}{N_i}||D_n^i(H_{out}^n) - D_n^i(H_{gt}^n)||_{1}}\right] \ ,
\end{split}
\end{equation}
where $L$ is the last layer of the discriminator, $D^i$ and $N_i$ are the activation map and its number of elements in the $i$-th layer of the discriminator, respectively. 

\textbf{Perceptual Loss.} To penalize the perceptual and semantic discrepancy, we employ the perceptual loss \cite{johnson2016perceptual} using a pertrained VGG-19 network:
\begin{equation}
\begin{split}
    \mathcal{L}_{perc} = \sum_i{\lambda_i||\Phi_i(I_{out}) - \Phi_i(I_{gt})||_{1}} \\
    +\lambda_l||\Phi_l(I_{out}) - \Phi_l(I_{gt})||_{2} \ ,
\end{split}
\end{equation}
where $\lambda_i$ are the balancing weights. $\Phi_i$ is the activation of $i$-th layer of the VGG-19 model which corresponds to the activation maps from layers \textit{relu1\_2}, \textit{relu2\_2}, \textit{relu3\_2}, \textit{relu4\_2} and \textit{relu5\_2}.
$\Phi_l$ represents the activation maps of \textit{relu4\_2} layer, and we select this specific layer to emphasize the high-level semantics. 

\textbf{Full Objective.}
With the linear combination of the aforementioned losses, the network is optimized by the following objective:
\begin{small}
\begin{equation}
    \mathcal{L}_\theta = \underset{G}{\min}\underset{D_{1},D_{2}}{\max}(\lambda_{l}\mathcal{L}_{LF}+\mathcal{L}_{G}+\lambda_{f}\mathcal{L}_{FM}+\lambda_{p}\mathcal{L}_{perc}) \ ,
\end{equation}
\end{small}
where we empirically set $\lambda_{l} = 2$, $\lambda_{f}=5$, and $\lambda_{p}=10$ in our experiments for balancing the objectives.

\section{Experiments}

\begin{figure*}[t]
\centering
\includegraphics[width=0.93\linewidth]{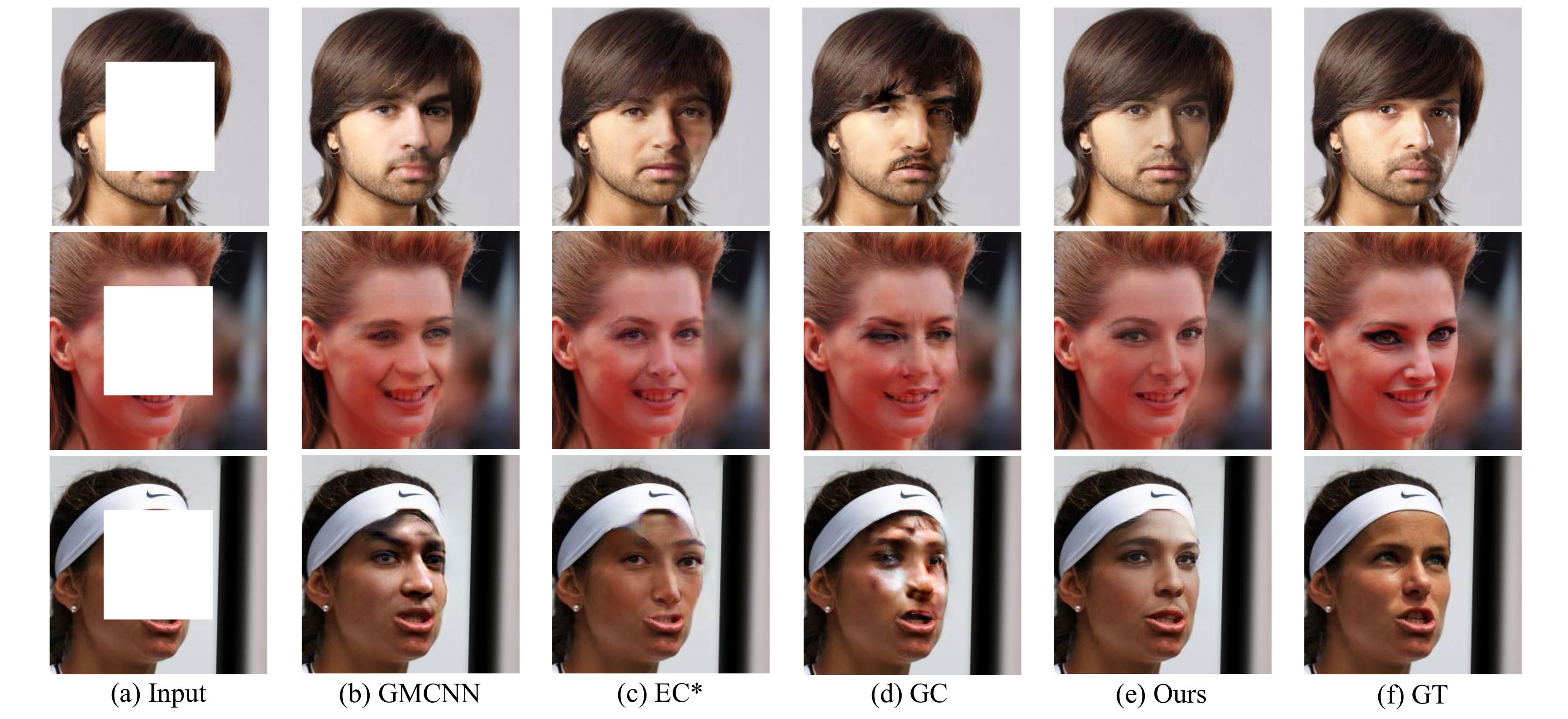}
\caption{
Qualitative comparison of WaveFill with the state-of-the-art: WaveFill generates more realistic inpainting with much less artifacts (over the dataset CelebA-HQ\cite{karras2017progressive} with central square masks). $\ast$ means that the model is trained with official implementation.}
\label{fig_comp_celebahq}
\end{figure*}

\subsection{Experimental Settings}
\textbf{Datasets.}
We conduct experiments on three public datasets that have different characteristics:
\begin{compactitem}
    \item[--] CelebA-HQ \cite{karras2017progressive}: It is a high-quality version of the human face dataset CelebA \cite{liu2015deep} with 30,000 aligned face images. We follow the split in \cite{yu2019free} that produces 28,000 training images and 2,000 validation images.

    \item[--] Places2 \cite{zhou2017places}: It consists of more than 1.8M natural images of 365 different scenes. We randomly sampled 10,000 images from the validation set in evaluations.

    \item[--] Paris StreetView \cite{pathak2016context}: It is a collection of street view images in Paris, which contains 14,900 training images and 100 validation images.
\end{compactitem}

\textbf{Compared Methods.}
We compare our method with a number of state-of-the-art methods as listed: 
\begin{compactitem}
    \item[--] GMCNN \cite{wang2018image}: It is a generative model with different receptive fields in different branches.
    \item[--] GC \cite{yu2019free}: It is also known as DeepFill v2, a two-stage method that leverages gated convolution.
    \item[--] EC \cite{nazeri2019edgeconnect}: It is a two-stage method that first predicts salient edges to guide the generation.  
    \item[--] MEDFE \cite{Liu2019MEDFE}: It is a mutual encoder-decoder that treats features from deep and shallow layers as structures and textures of an input image.
\end{compactitem}

\textbf{Evaluation Metrics.}
We perform evaluations by using four widely adopted evaluation metrics: 1) Fr{\'e}chet Inception Score (FID) \cite{heusel2017gans} that evaluates the perceptual quality by measuring the distribution distance between the synthesized images and real images; 2) mean $\ell_1$ error; 3) peak signal-to-noise ratio (PSNR); and 4) structural similarity index (SSIM) \cite{wang2004image} with a window size of 51.

\textbf{Implementation Details.}
The proposed method is implemented in PyTorch. The network is trained using $256 \times 256$ images with random rectangle masks or irregular masks \cite{liu2018image}. We use Adam optimizer \cite{kingma2014adam} with $\beta_{1}=0$ and $\beta_{2}=0.9$, and set the learning rate at 1e-4 and 4e-4 for the generator and discriminators, respectively. The experiments are conducted on 4 NVIDIA(R) Tesla(R) V100 GPU. The inference is performed in a single GPU, and our full model runs at 0.138 seconds per $256 \times 256$ image. 

\begin{figure*}[t]
\centering
\includegraphics[width=0.93\linewidth]{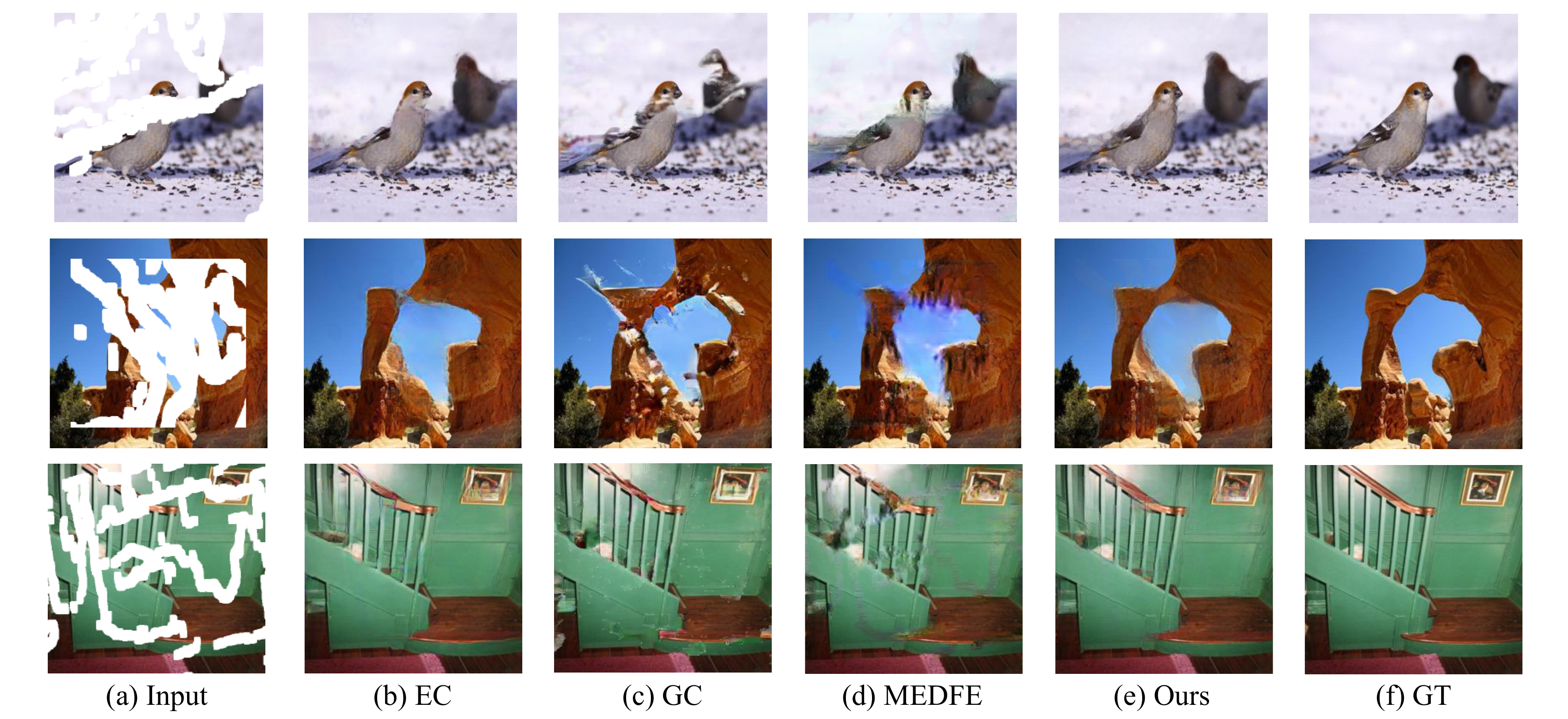}
\caption{
Qualitative comparison of WaveFill with the state-of-the-art: WaveFill generates more realistic inpainting with much less artifacts as compared with the state-of-the-art (over dataset Places2 \cite{zhou2017places} with irregular masks).
}
\label{fig_comp_places}
\end{figure*}

\renewcommand\arraystretch{1.2}
\begin{table}
\begin{center}
\begin{adjustbox}{width=\columnwidth,center}
\begin{tabular}{|m{1.2cm}<{\centering}|
                m{2.0cm}<{\centering}|
                m{1.5cm}<{\centering}|
                m{1.5cm}<{\centering}|
                m{1.5cm}<{\centering}|}
\hline
  & GMCNN \cite{wang2018image} &EC$\ast$ \cite{nazeri2019edgeconnect}  & GC \cite{yu2019free}   & \textbf{Ours} \\
\hhline{|=|=|=|=|=|}
\multirow{1}{*}{FID$\downarrow$} &8.17 &8.04 &7.39  &\textbf{6.48} \\ 
\hline
\multirow{1}{*}{$\ell_1(\%)\downarrow$} &2.38 &2.31 &2.53 & \textbf{2.26} \\ 
\hline
\multirow{1}{*}{PSNR$\uparrow$}  &25.86 &25.64 &25.37 &\textbf{26.53} \\ 
\hline
\multirow{1}{*}{SSIM$\uparrow$}  &0.905 &0.896 & 0.894& \textbf{0.911} \\ 
\hline
\end{tabular}
\end{adjustbox}
\end{center}
\caption{
Quantitative comparison of WaveFill with state-of-the-art methods over CelebA-HQ \cite{karras2017progressive} validation images (2,000) with square masks. $\ast$ denotes that we trained the model based on official implementation.
}
\label{tab_celebahq}
\end{table}

\subsection{Quantitative Evaluation}
We perform extensive quantitative evaluations over data with central square masks and irregular masks \cite{liu2018image}. For inpainting with central square masks, we use the mask size of $128\times{128}$, and benchmark with GMCNN \cite{wang2018image}, EC \cite{nazeri2019edgeconnect} and GC \cite{yu2019free} over the validation images of CelebA-HQ \cite{karras2017progressive}. For inpainting with irregular masks, we conducted experiments over Places2 \cite{zhou2017places} and Paris StreetView \cite{park2019semantic} and benchmarked with GC \cite{yu2019free}, EC \cite{nazeri2019edgeconnect} and MEDFE \cite{Liu2019MEDFE}. The irregular masks in the experiments are categorized based the ratios of the masked regions over the image size. Performance of the compared methods was acquired by running publicly available pre-trained.
The only exception is EC \cite{nazeri2019edgeconnect} which was trained with the official implementation on CelebA-HQ \cite{karras2017progressive} with random rectangle masks.

Table \ref{tab_celebahq} shows experimental results for dataset  CelebA-HQ with central square masks. It can be observed that WaveFill outperforms all existing methods under different evaluation metrics consistently. In addition, experiments with irregular masks show that WaveFill achieves superior inpainting under different mask ratios as shown in Table \ref{tab_places}. The effectiveness of WaveFill largely attributes to the wavelet-based frequency decomposition and the proposed normalization scheme. Specifically, disentangling frequency information in the wavelet domain helps mitigate the conflicts in generating low-frequency and high-frequency contents effectively, and it improves the inpainting quality in PSNR and SSIM as well. With the proposed normalization scheme, the low and high frequency information can be aligned for consistent generations in different frequency bands. Moreover, it allows the model to establish long-range dependencies which help generate more semantically plausible contents with better perceptual quality in FID. Quantitative results for Paris StreetView \cite{park2019semantic} are provided in the supplementary materials due to space limit.

\renewcommand\arraystretch{1.2}
\begin{table}
\begin{center}
\begin{adjustbox}{width=\columnwidth,center}
\begin{tabular}{|m{1.2cm}<{\centering}|
                m{1.5cm}<{\centering}|
                m{1.5cm}<{\centering}|
                m{1.5cm}<{\centering}|
                m{2cm}<{\centering}|
                m{1.5cm}<{\centering}|}
\hline
 & Mask & EC \cite{nazeri2019edgeconnect}  & GC \cite{yu2019free}  & MEDFE \cite{Liu2019MEDFE} & \textbf{Ours} \\
\hhline{|=|=|=|=|=|=|}
\multirow{4}{*}{FID$\downarrow$} & 10-20\% &2.55  &5.18  &2.81  &\textbf{1.96}  \\ \cline{2-6}
                                 & 20-30\% &5.36  &10.06 &7.51  &\textbf{4.08}  \\ \cline{2-6}
                                 & 30-40\% &9.28  &15.67 &15.84 &\textbf{7.33}  \\ \cline{2-6}
                                 & 40-50\% &15.17 &22.69 &28.98 &\textbf{12.68} \\ \cline{2-6}
\hhline{|=|=|=|=|=|=|}
\multirow{4}{*}{$\ell_1(\%)\downarrow$} & 10-20\% &1.55 &2.19 &1.42 &\textbf{1.39} \\ \cline{2-6}
                                        & 20-30\% &2.71 &3.73 &2.62 &\textbf{2.32} \\ \cline{2-6}
                                        & 30-40\% &3.97 &5.34 &4.13 &\textbf{3.42} \\ \cline{2-6}
                                        & 40-50\% &5.42 &7.05 &5.97 &\textbf{4.73} \\ \cline{2-6}
\hhline{|=|=|=|=|=|=|}
\multirow{4}{*}{PSNR$\uparrow$} & 10-20\% &27.23 &24.96 &28.48 &\textbf{28.72} \\ \cline{2-6}
                                & 20-30\% &24.30 &22.02 &24.76 &\textbf{25.87} \\ \cline{2-6}
                                & 30-40\% &22.31 &20.03 &22.05 &\textbf{23.74} \\ \cline{2-6}
                                & 40-50\% &20.67 &18.54 &19.87 &\textbf{21.99} \\ \cline{2-6}
\hhline{|=|=|=|=|=|=|}
\multirow{4}{*}{SSIM$\uparrow$} & 10-20\% &0.942 &0.906 &0.954 &\textbf{0.956} \\ \cline{2-6}
                                & 20-30\% &0.890 &0.833 &0.902 &\textbf{0.918} \\ \cline{2-6}
                                & 30-40\% &0.830 &0.758 &0.833 &\textbf{0.867} \\ \cline{2-6}
                                & 40-50\% &0.758 &0.679 &0.749 &\textbf{0.803} \\ \cline{2-6}

\hline
\end{tabular}
\end{adjustbox}
\end{center}
\caption{
Quantitative comparison of WaveFill with state-of-the-art methods over Places2 \cite{zhou2017places} validation images (10,000) with irregular masks \cite{liu2018image}.
}
\label{tab_places}
\end{table}
\subsection{Qualitative Evaluations}
Figs. \ref{fig_comp_celebahq} and \ref{fig_comp_places} show qualitative experimental results over the validation set of CelebA-HQ \cite{karras2017progressive} and Places2 \cite{zhou2017places}, respectively. As demonstrated in Fig. \ref{fig_comp_celebahq}, the inpainting by GMCNN \cite{wang2018image} and EC \cite{nazeri2019edgeconnect} suffers from unreasonable semantics and inconsistency near edge regions clearly, while the inpainting by GC \cite{yu2019free} contains obvious artifacts and blurry textures. As a comparison, the inpainting by WaveFill are more semantically reasonable and has less artifacts but more texture details. For dataset Places2 \cite{zhou2017places}, the inpainting by GC \cite{yu2019free} and MEDFE \cite{Liu2019MEDFE} contains undesired artifacts and distorted structures as shown in Figs. \ref{fig_comp_places}b and \ref{fig_comp_places}c. Though EC \cite{nazeri2019edgeconnect} produces more visually appealing contents with less artifacts, its generated semantics are still short of plausibility. Thanks to the frequency disentanglement and FRAN, WaveFill achieves superior inpainting for both central square masks and irregular masks.

\subsection{User Study}
\begin{figure}[t]
\centering
\includegraphics[width=1.0\linewidth]{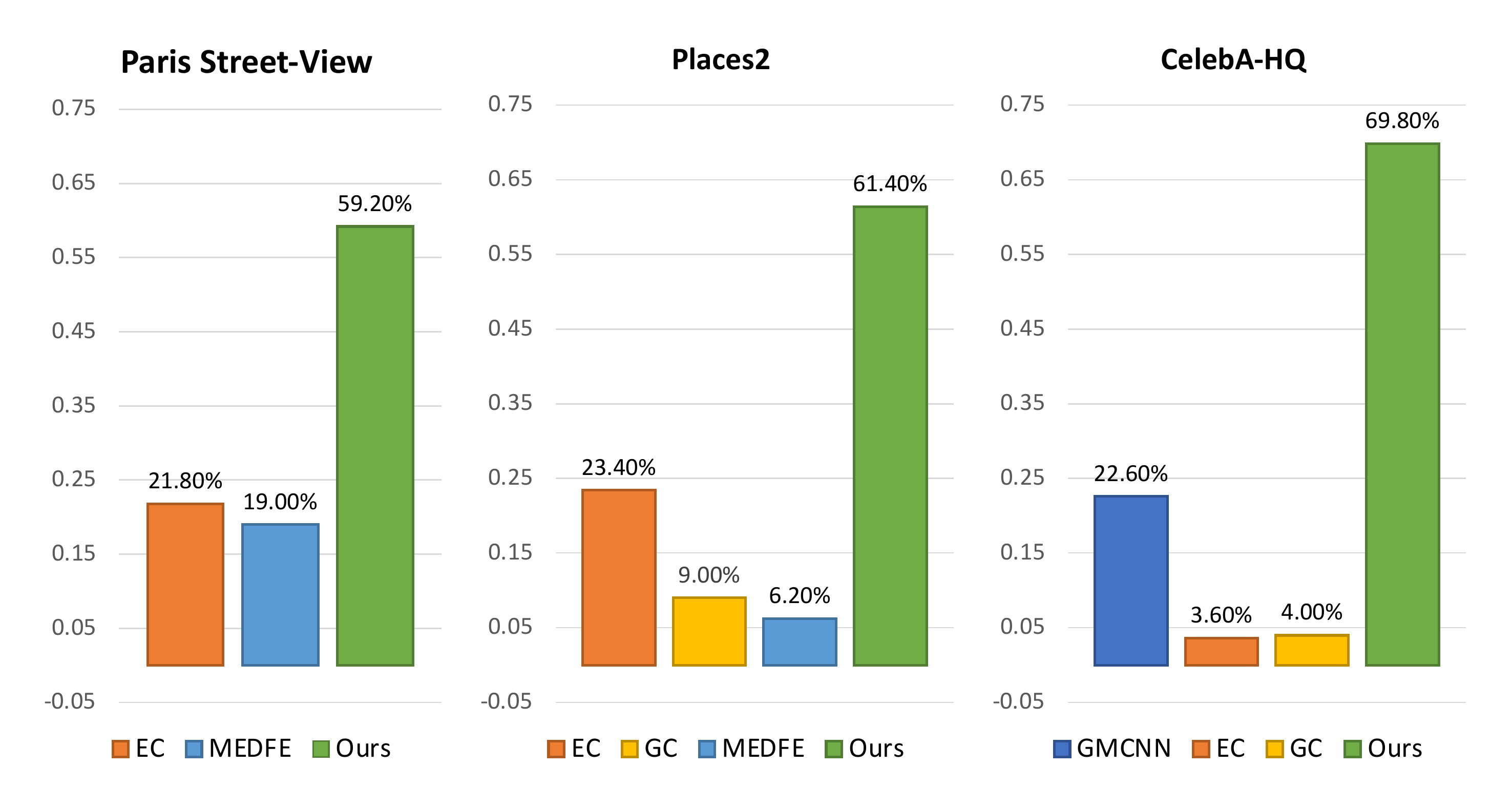}
\caption{
Inpainting evaluation by user study: The percentages tell the amount of images that are judged as the most realistic among images inpainted by all compared methods. 
}
\label{fig_us}
\end{figure}

We performed user studies over datasets Paris StreetView\cite{pathak2016context}, Places2\cite{zhou2017places} and CelebA-HQ\cite{karras2017progressive}. Specifically, we randomly sampled 25 test images from each dataset with no idea of inpainting results, which leads to 75 multiple choice questions in the survey. We recruited 20 volunteers with image processing backgrounds and each subject is asked to vote for the most realistic inpainting in each question. As Fig. \ref{fig_us} shows, the proposed WaveFill outperforms state-of-the-art methods by large margins.

\subsection{Ablation Study}

\renewcommand\arraystretch{1.2}
\begin{table}
\begin{center}
\begin{adjustbox}{width=\columnwidth,center}
\begin{tabular}{|l|cccc|}
\hline
Models & FID$\downarrow$ & $\ell_1(\%)\downarrow$ & PSNR$\uparrow$ & SSIM$\uparrow$ \\
\hline\hline
Spatial + Concat &33.95&2.45&28.37&0.898\\

DCT + Concat &100.93&4.71&23.56&0.765\\

Wavelet + Concat &32.73&2.46&28.46&0.899\\

Wavelet + SPADE  &32.14&2.38&28.81&0.901\\
\hline
Wavelet + FRAN  &\textbf{31.02}&\textbf{2.34}&\textbf{28.94}&\textbf{0.904}\\
\hline
\end{tabular}
\end{adjustbox}
\end{center}
\caption{
Ablation study of WaveFill over Paris StreetView \cite{park2019semantic} validation set (100) with irregular masks \cite{liu2018image}. Model in the last row is the standard WaveFill.} 
\label{tab_as1}
\end{table}



We study the individual contributions of our technical designs by several ablation studies over Paris StreetView \cite{park2019semantic} as shown in Table \ref{tab_as1}. In the ablation studies, we trained four network models including: 1) \textit{Spatial + Concat} (Baseline) that adopts the typical encoder-decoder network with gated convolution \cite{yu2019free}. Different from WaveFill, L1 and adversarial losses are applied together, multi-level features are directly concatenated; 2) \textit{DCT + Concat} that adopts discrete cosine transform (DCT) to compare with wavelet transformation. Similar to WaveFill, we split the frequency bands into three groups and feed them to the three generation branches; 3) \textit{Wavelet + Concat} that replaces FRAN by concatenation of multi-frequency features; 4) \textit{Wavelet + SPADE } that replace FRAN by SPADE \cite{park2019semantic}.

\begin{figure}[t]
\centering
\includegraphics[width=1.0\linewidth]{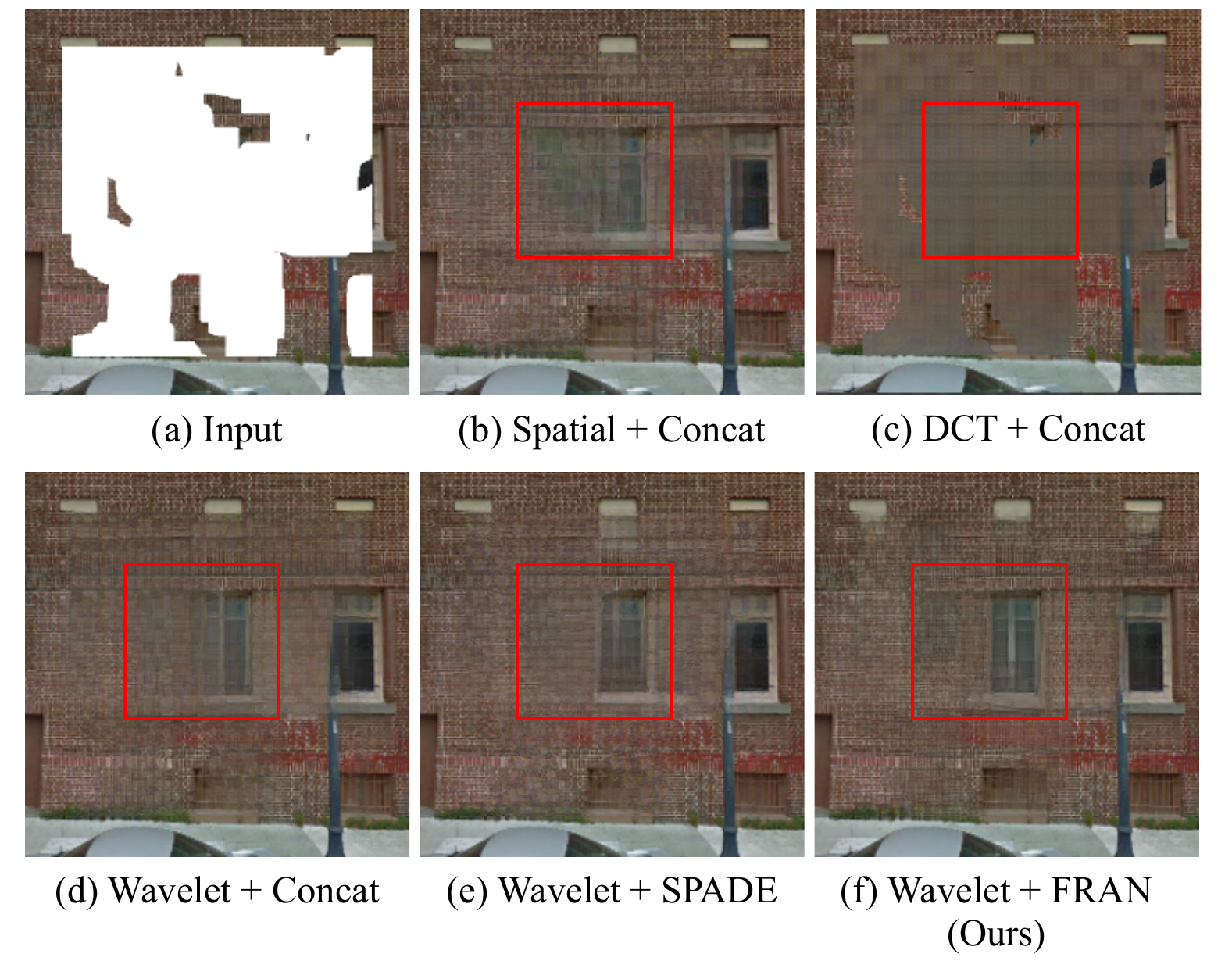}
\caption{
Ablation study of the proposed WaveFill: Our `Wavelet’ and `FRAN’ designs both help suppress artifacts and synthesize plausible semantics effectively. The study is performed over Paris StreeView \cite{pathak2016context} with irregular masks. The red boxes are used to highlight the main differences across different approaches.
}
\label{fig_comp_as}

\end{figure}

As shown in Table \ref{tab_as1}, using DCT degrades the inpainting greatly due to the lack of spatial information. Wavelet transformation preserves spatial information which improves inpainting by large margins. In addition, using wavelet outperforms the baseline especially in FID, largely because wavelet-based model disentangles multi-frequency information and recovers corrupted regions in different frequency bands separately. Visual evaluation is well aligned with quantitative experiments in Fig. \ref{fig_comp_as}. We can see that DCT-based model fails to synthesize meaningful structures as shown in (c). Spatial-based model instead introduces unreasonable semantics and clear artifacts as shown in (b). Our wavelet-based model fills the missing regions with much less artifacts as shown in (d). Further, concatenation and SPADE do not align the features of different frequencies for better content consistency. FRAN addresses this issue effectively as shown in Table \ref{tab_as1} and Fig. \ref{fig_comp_as}.
More ablation studies are included in the supplementary materials.

\section{Conclusion}

This paper presents WaveFill, a novel image inpainting framework that disentangles low and high frequency information in the wavelet domain and fills the corrupted regions explicitly and separately. To ensure the inpainting consistency across multiple frequency bands, we propose a novel frequency region attentive normalization (FRAN) that effectively aligns and fuses the multi-frequency features especially those within the missing regions. Extensive experiments show that WaveFill achieves superior image inpainting for both rectangle and free-form masks. 
Moving forward, we will study how to adapt the idea of wavelet decomposition and separate processing in different frequency bands to other image recovery and generation tasks.

{\small
\bibliographystyle{ieee_fullname}
\bibliography{egbib}
}

\end{document}